\documentclass[10pt,twocolumn,letterpaper]{article}

\usepackage{iccv}
\usepackage{times}
\usepackage{epsfig}
\usepackage{graphicx}
\usepackage{amsmath}
\usepackage{amssymb}

\usepackage[breaklinks=true,bookmarks=false]{hyperref}

\iccvfinalcopy 

\def\iccvPaperID{****} 

\begin{document}

\title{3D Dense Face Alignment via Graph Convolution Networks}

\author{Huawei Wei, Shuang Liang, Yichen Wei\\
Megvii Technology, Tongji University\\
\{weihuawei, weiyichen\}@megvii.com\\
shuangliang@tongji.edu.cn
}

\maketitle

\begin{abstract}
Recently, 3D face reconstruction and face alignment tasks are gradually combined into one task: 3D dense face alignment. Its goal is to reconstruct the 3D geometric structure of face with pose information. In this paper, we propose a graph convolution network to regress 3D face coordinates. Our method directly performs feature learning on the 3D face mesh, where the geometric structure and details are well preserved. Extensive experiments show that our approach gains a superior performance over state-of-the-art methods on several challenging datasets.
\end{abstract}

\section{Introduction}
3D face reconstruction and face alignment have been widely used in film and animation production. The two tasks are highly related and well studied. Originally, 3D face reconstruction~\cite{facerecon1,facerecon2} aims to recover the 3D face geometry and face alignment aims to locate a number of fiducial facial landmarks~\cite{2dface1,2dface2}. Recently, a number of works have integrated them into a single task, so called \emph{3D dense face alignment}. It aims to recover the 3D face geometry as well as its pose.

\begin{figure}[t]
\centering
\includegraphics[width=3in]{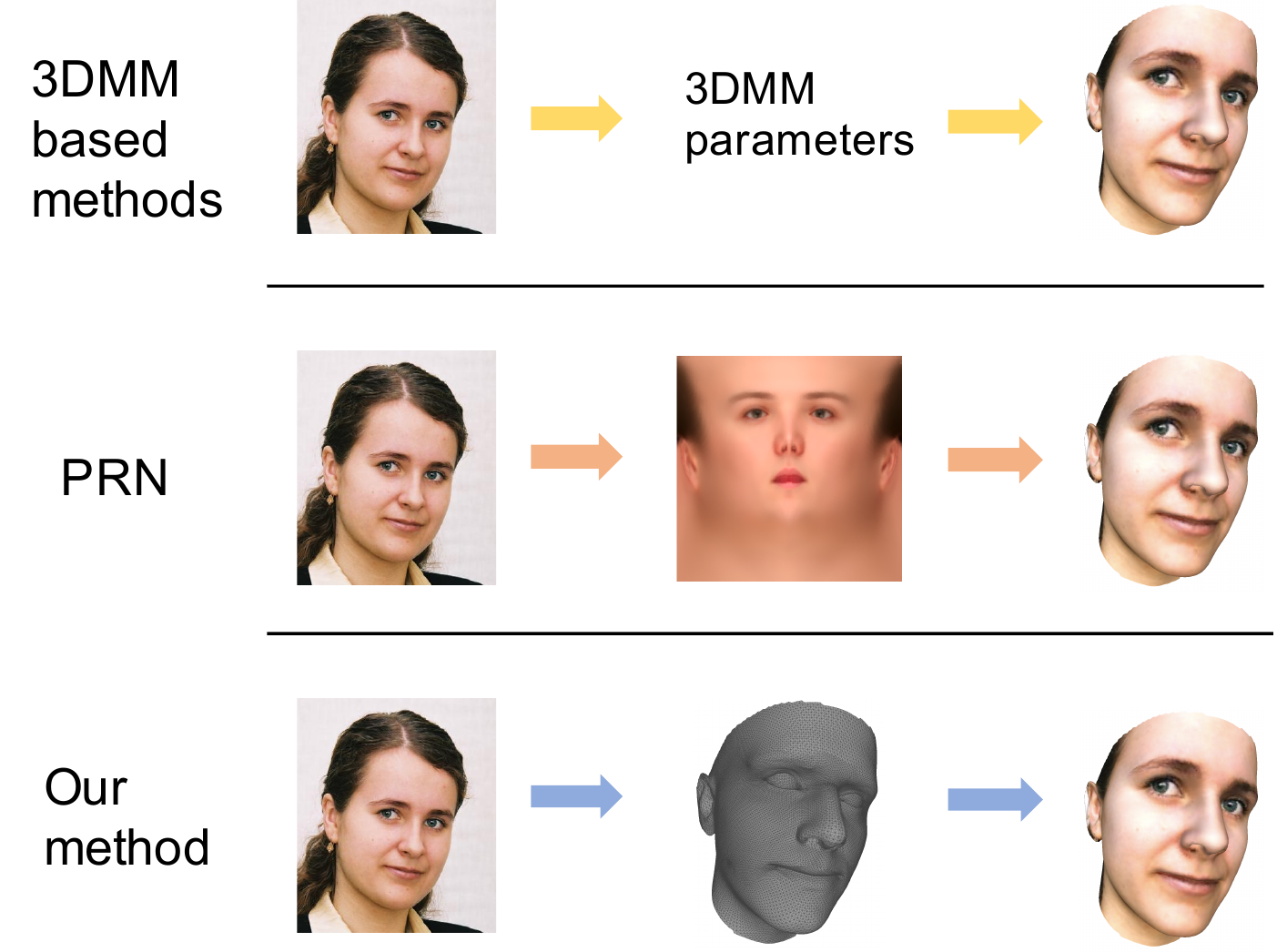}
\caption{Illustration of three different 3D dense face alignment methods. Top row: 3DMM based methods~\cite{3ddfa, defa}. Middle row: PRN~\cite{prn} uses uv position map to store 3D face coordinates. Bottom row: our graph convolution method directly regresses 3D face coordinates on the mesh.}
\label{fig:example_of_3d_methods}
\end{figure}

Previous 3D dense face alignment methods fall into two categories. The \emph{first} category fits a parametric model, specifically, the coefficients of a 3D Morphable Model (3DMM)~\cite{3dmm} and the projection matrix from a 2D face image~\cite{3ddfa, defa}. The recovered geometry accuracy is up to the capacity of the 3D Morphable Model. Usually, it is hard to recover the details that are missing in 3DMM.

The \emph{second} category directly estimates the 3D face coordinates in grid-like data structure (e.g., volumetric representation~\cite{vrn} and uv position map~\cite{prn}), so that convolutional neutral networks (CNNs) can be used for the regression. Such methods are capable of recovering the geometric details. However, their grid-like data structure is misaligned with the 3D geometry data format (usually a mesh). This causes errors by \emph{data representation}. For example, volumetric representation~\cite{vrn} introduces quantization errors during voxelization of the mesh. The generated 3D face is usually grainy. uv position map~\cite{prn} introduces distortion, especially in the exterior area of the position map. This is illustrated in Fig.~\ref{fig:example_of_3d_methods} (middle).

To address the problem in the second category, we propose to directly regress the 3D coordinates on the face mesh, as shown in Fig.~\ref{fig:example_of_3d_methods}. Because 3D face mesh is represented by a graph, ordinary CNNs are not suitable to process it. Although one can use fully connected network for such regression, this is brute force and computationally unaffordable. Instead, we propose to use graph convolution networks~\cite{chebynet,gcn} for the task, which are born to work with the graph data. Our method directly performs feature learning on the 3D face mesh, where the geometric structure is well preserved. There is no loss in data representation as in~\cite{prn,vrn}. 3D geometric details are recovered. 

The contribution of this work is two fold. First, \emph{for the first time}, we propose to use graph convolution for the 3D dense face alignment task and develop a novel implementation. We integrate well developed techniques in standard CNNs, such as encoder-decoder architecture, residual learning~\cite{resnet} and Instance Normalization~\cite{in} with graph convolution. A coarse to fine strategy is developed using mesh sampling techniques~\cite{coma,quadric}. The resulting network architecture converges well in training and is fast in inference.

Second, comprehensive experiments are performed on several challenging datasets. Extensive quantitative and qualitative results show that our approach gains better results than state-of-the-art methods. This verifies the effect of our approach.

\section{Related work}

\textbf{3D dense face alignment} An early representative work is 3DDFA~\cite{3ddfa}. It uses a cascaded CNN to fit the 3DMM~\cite{3dmm} parameters and the projection matrix from a single 2D face image. It shows promising alignment result across large poses, and defines the new problem called 3D dense face alignment. ~\cite{defa} utilizes multi-constraints such as face contours and SIFT feature points to estimate the 3D face shape. It provides a very dense 3D alignment. Although these 3DMM based methods achieve good performance, their accuracy is up to the capacity of the 3DMM. PRN~\cite{prn} uses a simple encoder-decoder architecture to regress the 3D face coordinates, which are stored in uv position map. Although it achieves state-of-the-art performance, there are many stripes on the reconstructed 3D face. This is attributed to the geometry distortion caused by uv mapping.

\textbf{Graph convolution networks} CNNs are suitable to process grid-like data such as images. By contrast, graph convolution networks (GCNs) are suitable to process graph data such as 3D mesh. A comprehensive overview of GCNs is provided by~\cite{geometric}. ~\cite{meshgcn1} define the first graph convolution operator on meshes by parameterizing the surface around each point using geodesic polar coordinates and performing graph convolution on the angular bins. Then, different parameterization methods are proposed by ~\cite{meshgcn2,meshgcn3}, but the manners of graph convolution are similar. These methods only present generalization of convolutions to
meshes. They do not design sampling operation on meshes. Therefore, coarse-to-fine features cannot be captured.

\cite{spectral} develops the spectral graph convolution in Fourier space. However, it is computationally expensive and unable to obtain the local features on the graph. To address these problems, ChebyNet~\cite{chebynet} formulates spectral convolution as a recursive Chebyshev polynomial, which avoids computing the Fourier basis. Recently, CoMA~\cite{coma} extends ChebyNet to process 3D meshes. It constructs an autoencoder to learn a latent representation of 3D face and introduces a
mesh pooling operator. By utilizing the spectral graph convolution and mesh sampling operations, CoMA~\cite{coma} obtains state-of-the-art results in 3D face modeling. Motivated by CoMA~\cite{coma}, for the first time, this work uses graph convolution for 3D dense face alignment.

\section{Method}
\subsection{Graph convolution on face mesh}
We briefly review spectral graph convolution, applied on face mesh. More details can be referred to~\cite{chebynet}.

A 3D face mesh is defined as $\mathcal{M}=(\mathcal{V}, W)$. $\mathcal{V}$ has $N$ 3D vertices on the 3D face surface, $\mathcal{V} \in \mathbb{R}^{N \times 3}$. $W$ is a sparse adjacency matrix of $\mathcal{V}$, $W \in\{0,1\}^{N \times N}$. $W_{ij}=1$ denotes there is an edge between vertices $i$ and $j$, and $W_{ij}=0$ otherwise. The normalized Laplacian matrix is $L = I -D^{-\frac{1}{2}}WD^{-\frac{1}{2}}$, where $I$ is identity matrix and $D$ is a diagonal matrix with $D_{i, i}=\sum_{j=1}^{n} W_{i, j}$. The spectral graph convolution of $x$ and $y$ is defined as a Hadamard product in the Fourier space, $x * y=U\left(\left(U^{T} x\right) \odot\left(U^{T} y\right)\right)$. $U$ is the eigenvectors of Laplacian matrix~\cite{laplacian}. Since $U$ is not sparse, this operation is computationally expensive. To reduce computation, \cite{chebynet} formulates spectral convolution with a kernel $g_\theta$ using a recursive Chebyshev polynomial, denoted as:
\begin{gather}
g_{\theta}(L)=\sum_{k=0}^{K-1} \theta_{k} T_{k}(\tilde{L}),
\end{gather}
where $\tilde{L}=2 L / \lambda_{\max }-I$ is the scaled Laplacian matrix, $ \lambda_{\max }$ is the maximum eigenvalue of the Laplacian matrix, $\theta \in \mathbb{R}^{K}$ is the Chebyshev coefficients, and $T_k$ is the Chebyshev polynomial of order $k$, which is computed recursively as $T_{k}(x)=2 x T_{k-1}(x)-T_{k-2}(x)$ with $T_0 =1$ and $T_1 =x$. 

For each layer, the spectral graph convolution is
\begin{gather}
h_{j}=\sum_{i=1}^{F_{i n}} g_{\theta_{i, j}}(L) x_{i},
\end{gather}
where $x_i$ is the $i$-th feature of input $x \in \mathbb{R}^{N \times F_{in}}$, $h_j$ is the $j$-th feature of output $h \in \mathbb{R}^{N \times F_{o u t}}$. There are $F_{i n} \times F_{o u t}$ vectors of Chebyshev coefficients in a convolution layer.

\begin{figure}[t]
\centering
\includegraphics[width=3in]{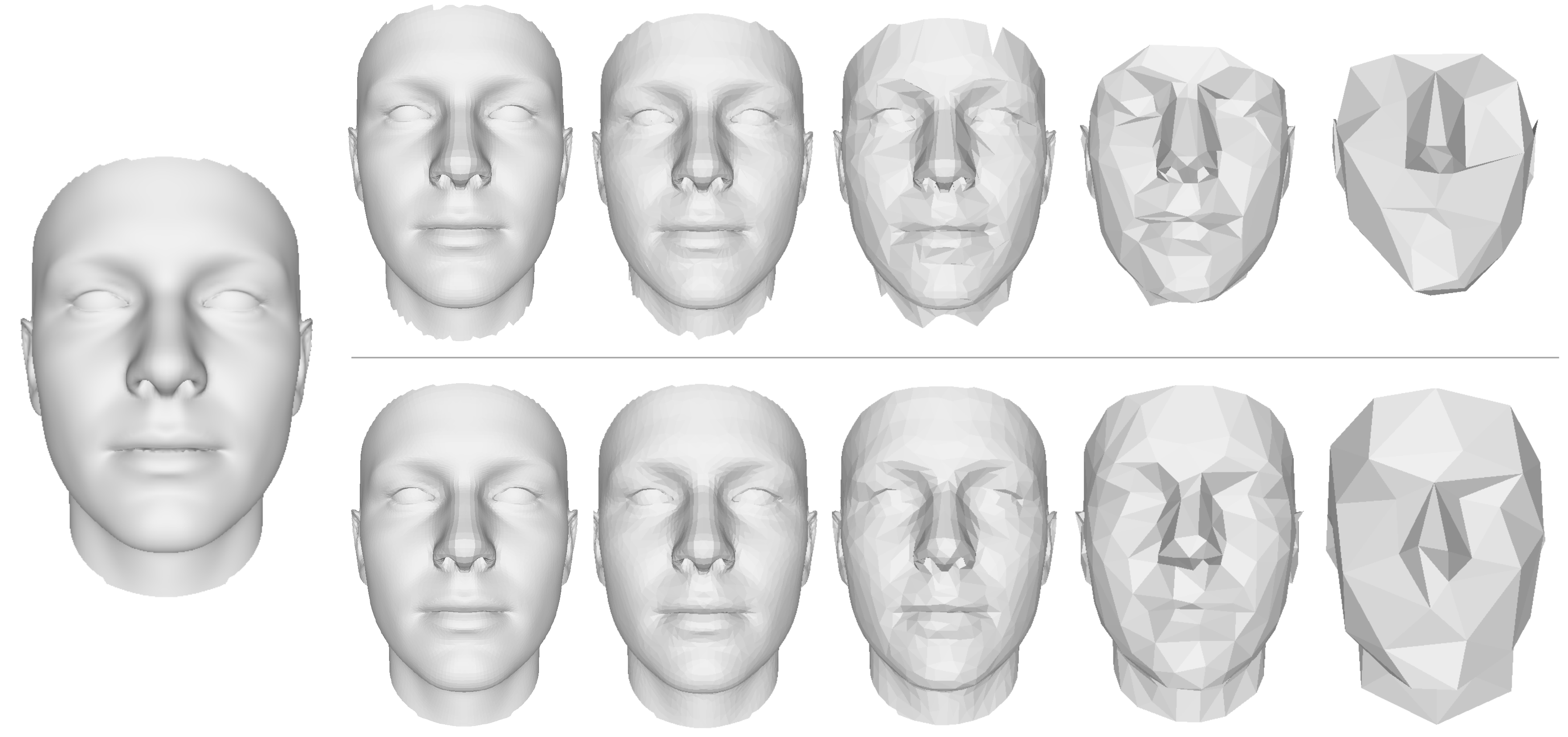}
\caption{On the leftmost is the 3DMM~\cite{3dmm} mesh used in this work. It has 105,954 triangle faces and 53,215 vertices. On the right are down sampled meshes, with 26356, 6528, 1599, 382, 84 triangle faces and 13304, 3326, 832, 208, 52 vertices, respectively. Top/bottom row: without/with boundary preservation~\cite{quadric}.}
\label{fig:face_mesh_sampling}
\end{figure}

\subsection{Face mesh sampling}
We use the 3DMM~\cite{3dmm} face mesh. It is shown in Fig.~\ref{fig:face_mesh_sampling}. Directly working on the high resolution in the original mesh is computationally prohibitive. As in standard CNNs, we develop a coarse-to-fine feature representation. The face mesh is down sampled to several different resolutions, using quadric error metrics~\cite{quadric}. Results are illustrated in Fig.~\ref{fig:face_mesh_sampling}. The idea is to minimize the quadric error between the simplified and the original meshes. Since 3D face mesh has open boundaries, performing simplification operation directly as did in~\cite{coma} causes too much distortion. Therefore, we add boundary preservation constraints as in~\cite{quadric} to preserve the geometry fidelity. As shown in Fig.~\ref{fig:face_mesh_sampling}, adding boundary preservation constraints effectively preserves the face geometry. This improves the feature learning in the subsequent graph convolution networks.

During the down-sampling procedure, we follow the method in~\cite{coma} to compute the up-sampling matrix. More details can be found in~\cite{coma}. It is used to restore the mesh resolution in the coarse-to-fine feature learning.

\begin{figure*}[t]
\centering
\includegraphics[width=7in]{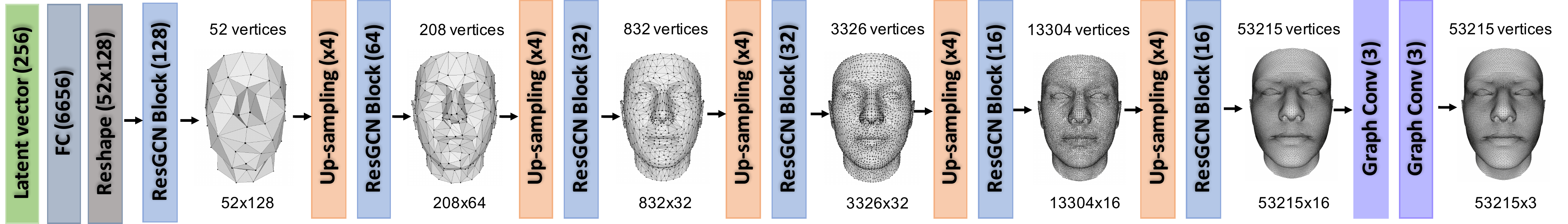}
\caption{The structure of the decoder. It consists of 6 graph convolution residual blocks, corresponding to the 6 resolution levels of the face mesh in Fig.~\ref{fig:face_mesh_sampling}. Each block is followed by an up-sampling operation except for the last one. The last graph convolution layer generates the 3D face vertices' coordinates.}
\label{fig:decoder}
\end{figure*}

\begin{figure}[t]
\centering
\includegraphics[width=1.6in, height=2.5in]{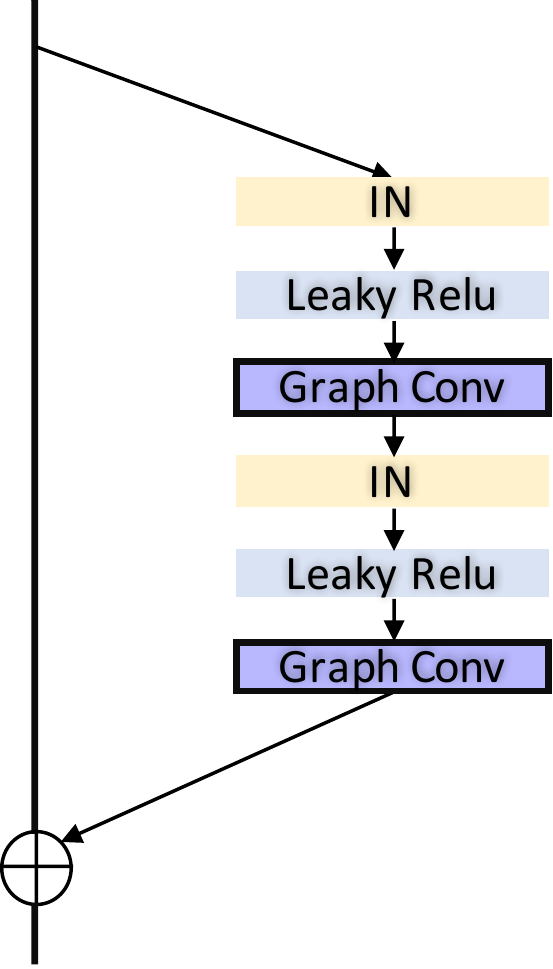}
\caption{Our proposed graph convolution residual block.}
\label{fig:ResGCN_Block}
\end{figure}

\subsection{Our proposed network}
Our proposed network consists of an encoder and a decoder. The encoder is Resnet-50~\cite{resnet}. It encodes the input 2D face image into a feature vector. The decoder is shown in Fig.~\ref{fig:decoder}. It restores the 3D face mesh vertices from the encoded feature in a hierarchical, coarse-to-fine manner. It consists of 6 graph convolution residual blocks, called ResGCN Block in this work, and two graph convolution layers. Each ResGCN block corresponds to a mesh resolution in Fig.~\ref{fig:face_mesh_sampling} and performs feature learning on that level. It is followed by an up-sampling operation, except for the last one. 

The ResGCN Block is illustrated in Fig.~\ref{fig:ResGCN_Block}. The identity path in the block helps convergence in the training. Instead of the commonly used Batch Normalization (BN)~\cite{bn}, Instance Normalization (IN)~\cite{in} is used in the block. This is because there is no strong statistical connection between face meshes under different poses. On the contrary, there is strong correlation among coordinates of a single face mesh since their combination determines the orientation of the face. Therefore, IN is preferred over BN. Leaky Relu~\cite{leakyrelu} is the activation function. Each block contains two graph convolution layers. The channel numbers of the six ResGCN Block are 128, 64, 32, 32, 16 and 16 respectively. The output features from the last ResGCN block are fed in the last two graph convolution layers to generate the 3D face vertices' coordinates.

In experiments, we found the network converges well in training. If the identity path in the block is removed, or BN is used instead of IN, the training does not converge.

\paragraph{Loss function}
Similar to~\cite{coma}, we adopt $\mathcal{L}_1$ loss for predicted 3D face vertices' coordinates $\tilde{Y}$. For each 2D face image, its ground truth 3D mesh $\mathcal{M}$ is obtained by Basel Face Model(BFM)~\cite{3dmm}.

We also use a smooth loss $\mathcal{L}_{smooth}$, denoted as:
\begin{gather}
\mathcal{L}_{smooth}=||(D-W)\tilde{Y}||_2.
\end{gather}
Note that $(D-W)$ is the unnormalized Laplacian matrix~\cite{laplacian} of mesh $\mathcal{M}$. Thus, $(D-W)\tilde{Y}$ denotes the difference between each vertex and its surrounding vertices. Adding the smoothness loss regularizes the training. A weight $\alpha$ is used to balance the smoothness and $\mathcal{L}_1$.

\begin{figure*}[t]
\centering
\includegraphics[width=6.5in]{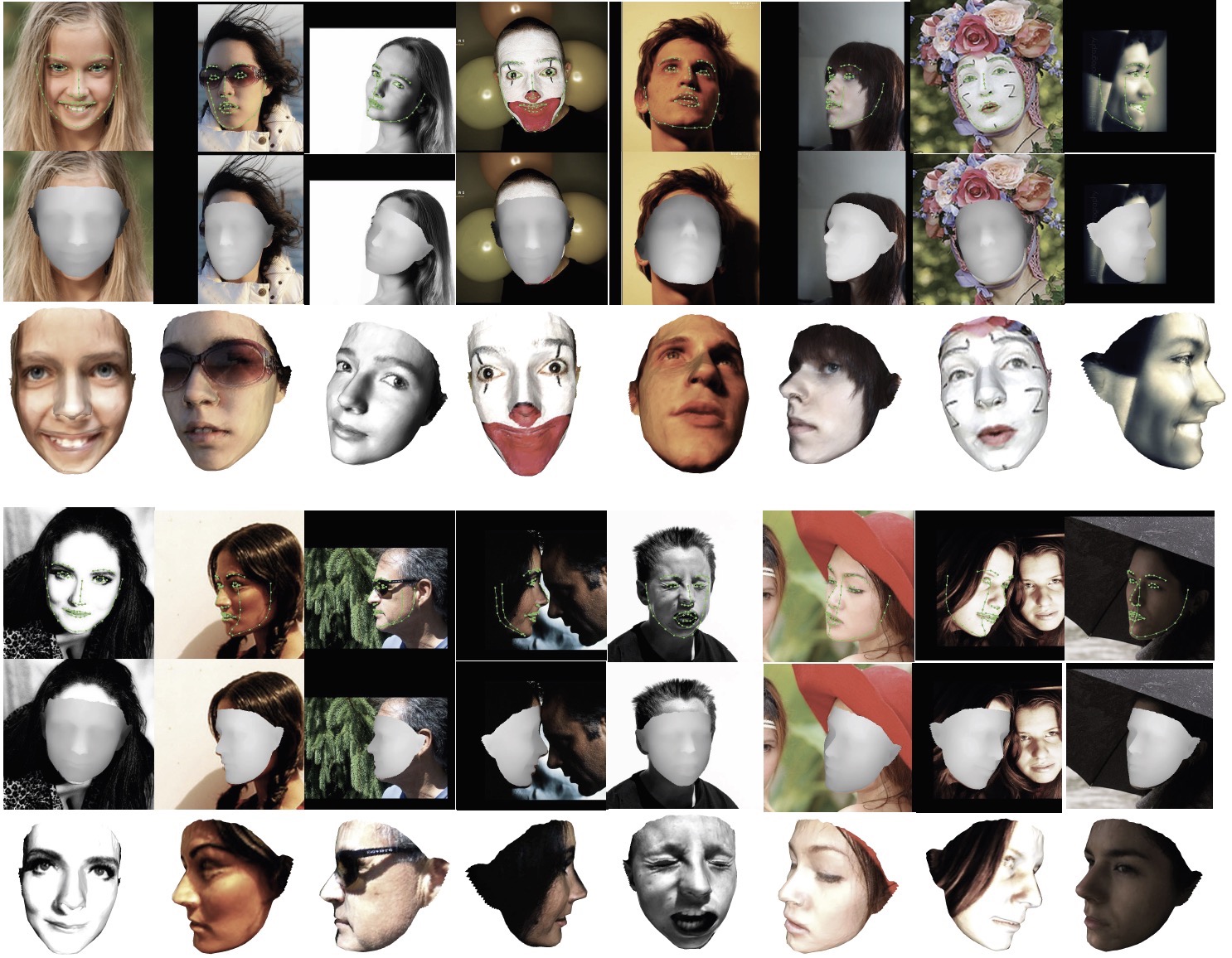}
\caption{The qualitative results of our 3D dense face alignment method. The first row of each person is the alignment results(68 landmarks are plotted), the second row is the face rendered by the corresponding depth image, the last row is the 3D reconstruction results.}
\label{fig:mask_display}
\end{figure*}

\begin{figure*}[t]
\centering
\includegraphics[width=6.5in]{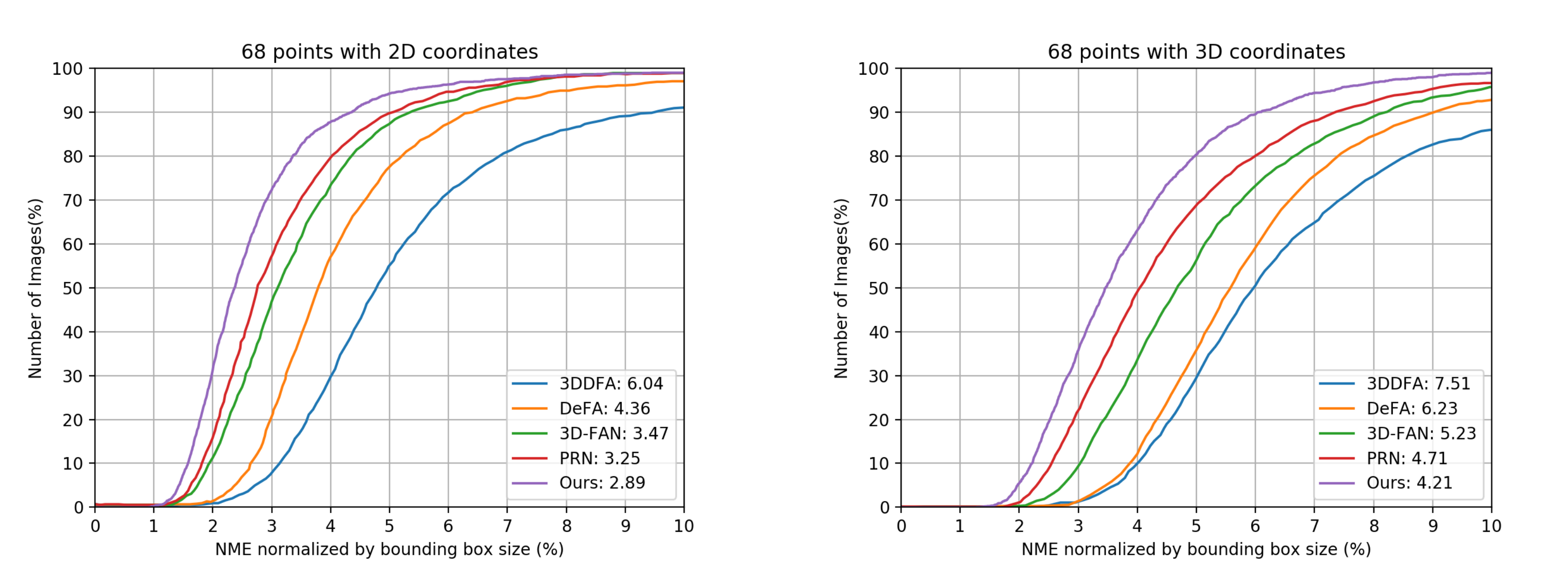}
\caption{Cumulative Errors Distribution (CED) curves of 68 points face alignment on AFLW2000-3D. The left is result of 2D face alignment. The right is the result of 3D face alignment.}
\label{fig:5}
\end{figure*}

\section{Experiments}
\subsection{Training}
We use 300W-LP~\cite{3ddfa} as our training dataset, the same as ~\cite{prn}. 300W-LP contains more than 60K unconstrained images with fitted 3DMM parameters and pose coefficients. The 3DMM parameters are obtained by Basel Face Model (BFM)~\cite{3dmm}. For each 2D face image in 300W-LP~\cite{3ddfa}, we first use BFM to generate the corresponding 3D mesh. The pose coefficients are then utilized to transform it as the ground truth. We compute the bounding box using the labeled 68 facial keypoints provided by 300W-LP, and then use the the bounding box to crop the image and resize the cropped image to size 256$\times$256. Following the same setting as in~\cite{prn}, we randomly rotate the input image by -45 to 45 degrees and perturb it with a random translation of 10\% of the input size. In addition, a random scale from 0.85 to 1.15 is added. The order number $K$ of all graph convolution layers is set as 3. The dimension of the encoding vector is 256. We train our network for 80 epochs with batch size 50. We choose Adam optimizer~\cite{adam}, where the initial learning rate is 0.001, decayed by half every 20 epochs. The weight $\alpha$ of smooth loss is 0.1. All experiments are conducted on the Geforce GTX 1080 Ti GPU using Tensorflow~\cite{tf}.

\subsection{Evaluation datasets and metrics}
To verify the performance on both face alignment and face reconstruction tasks, we choose the following three datasets as our evaluation benchmarks.

\textbf{AFLW2000-3D}~\cite{3ddfa} contains the first 2000 images from AFLW~\cite{aflw} whose annotations include 68 3D facial landmarks and the fitted 3DMM parameters. Performance of our method on face alignment and face reconstruction tasks is evaluated on this dataset.

\textbf{AFLW-LFPA} is constructed by~\cite{aflwp}. It contains 1299 images with a balanced distribution of face postures. For each image, 34 facial landmarks are provided. This database is used for evaluation on face alignment task.

\textbf{Florence}~\cite{florence} is a common used benchmark for 3D face reconstruction. It consists of high-resolution 3D scans of 53 subjects. We follow the protocol of~\cite{vrn} to generate renderings with different poses, using a pitch of -15, 20 or 25 degrees and each of the five
evenly spaced rotations between -80 and 80 degrees.

We employ the Normalized Mean Error(NME) as the evaluation metric, which is:

\begin{gather}
\mathrm{NME}=\frac{1}{N} \sum_{i=1}^{N} \frac{\left\|Y_{i}-\tilde{Y}_{i}\right\|_{2}}{d},
\end{gather}

where $Y_i$ is the $i$-th coordinate of the ground truth 3D face and $\tilde{Y}_i$ denotes the $i$-th coordinate of the predicted 3D face. $d$ is the normalization factor.

\subsection{Results}
In this part, we first present the 2D and 3D face alignment performance compared with several state-of-the-art methods on AFLW2000-3D and AFLW-LFPA datasets. Then the results of 3D face reconstruction on Florence are shown. At last, ablation study about our different experimental setting is demonstrated. The qualitative results are shown in Fig.~\ref{fig:mask_display}, notice that our method can guarantee good face alignment and 3D face reconstruction performance even in cases of large pose, occlusion and weak illumination. 

\begin{figure*}[t]
\centering
\includegraphics[width=6.5in]{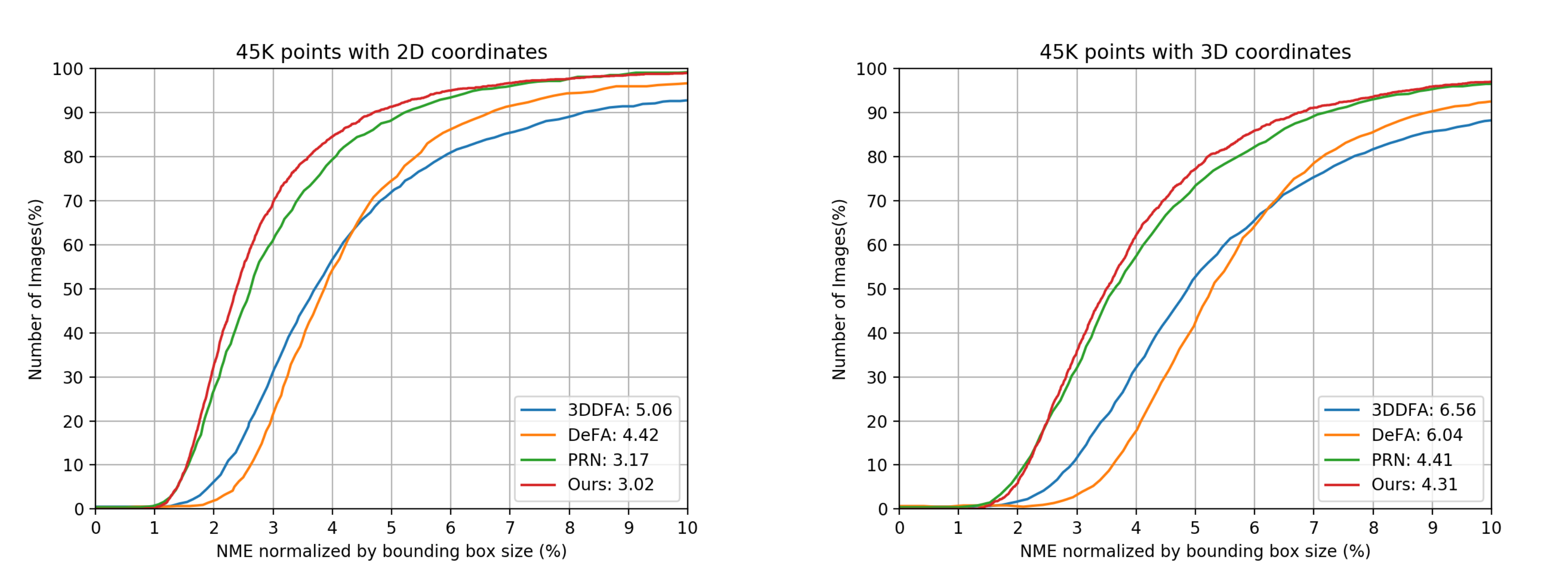}
\caption{Cumulative Errors Distribution (CED) curves of 45K points face aligment on AFLW2000-3D. The left is result of 2D face alignment. The right is the result of 3D face alignment.}
\label{fig:6}
\end{figure*}

\subsubsection{Face alignment}
We first conduct face alignment experiments on 68 \emph{sparse} facial landmarks on AFLW2000-3D. We follow~\cite{3ddfa} to use bounding box size as the normalization factor. Several recent state-of-the-art methods are selected for comparison, including 3DDFA~\cite{3ddfa}, DeFA~\cite{defa}, 3D-FAN~\cite{3dfan} and PRN~\cite{prn}. As shown in Fig.~\ref{fig:5}, our approach outperforms other methods both on 2D and 3D face alignment tasks by a large margin. Specifically, more than 10\% higher performance is achieved compared with the best 3D face alignment method. It shows that our approach can locate landmarks more accurately.

\begin{figure*}[t]
\centering
\includegraphics[width=6in]{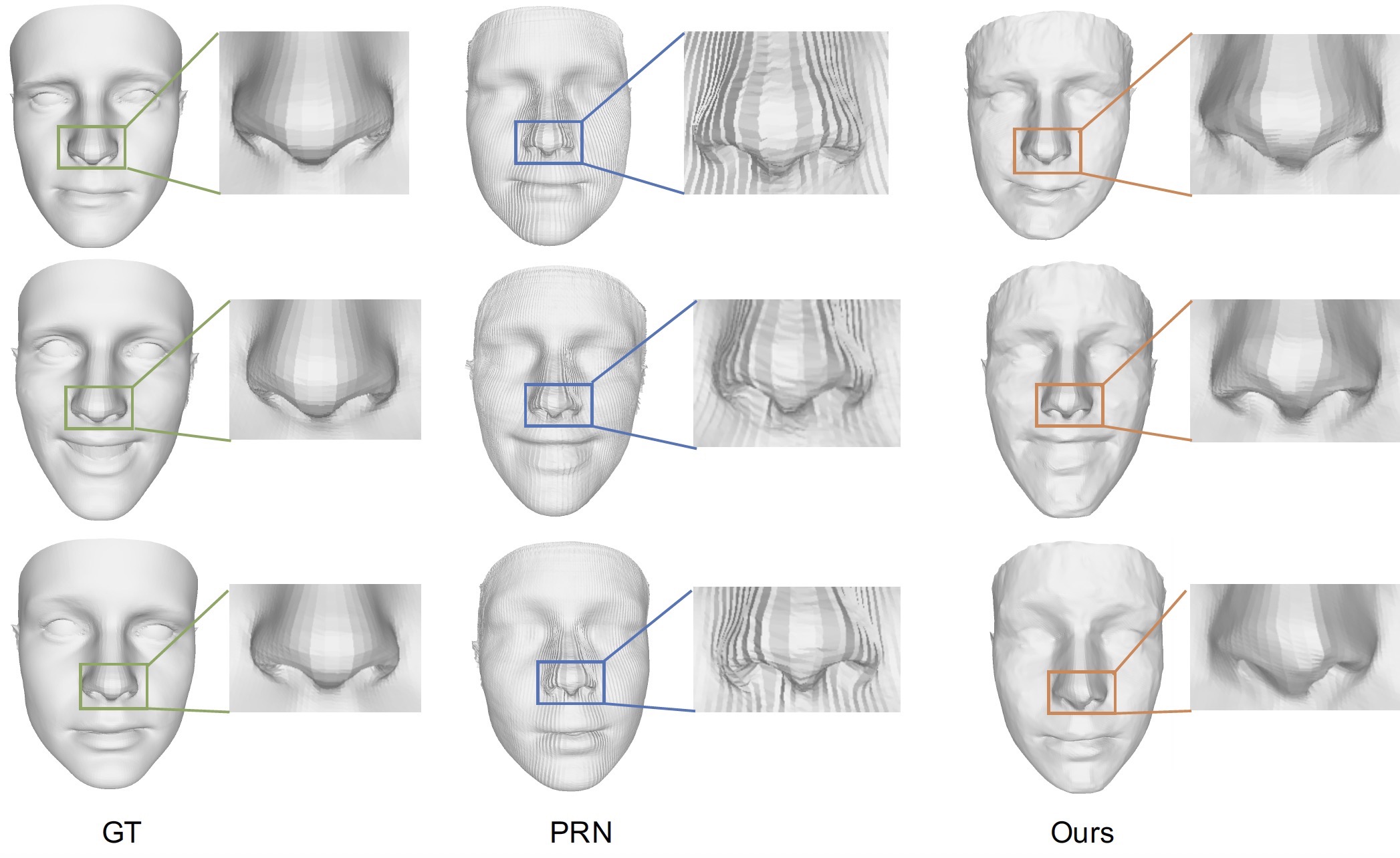}
\caption{Examples of the reconstructed 3D faces. The left is the ground truth, the middle is result of PRN and the right is our result. The nose area is zoomed in for better view.}
\label{fig:prn_and_ours_result}
\end{figure*}

In addition, we also present our \emph{dense} face alignment results compared with other recent methods including 3DDFA~\cite{3ddfa}, DeFA~\cite{defa} and PRN~\cite{prn} on AFLW2000-3D. Follow the setting of PRN, we select 45K points from the largest common face region of all compared methods. The quantitative results is illustrated in Fig.~\ref{fig:6}. Our method is superior to the best state-of-the-art method. In addition, our generated 3d faces have better visual quality. Examples and comparison are illustrated in Fig.~\ref{fig:prn_and_ours_result}. For better display, we rotate the 3d mesh into a front face and zoom in the nose area. Note that there are many stripes on face surface of PRN. By contrast, our results are smoother. Besides, our generated face has finer details and better correspondence to the ground truth. By comparison, PRN adopts uv position map as the regression target, there is geometric distortion between it and the 3D coordinates. Therefore, it produces less precise results.

To investigate the performance of our method across different poses and datasets, we conduct face alignment experiments with different yaw angles on AFLW2000-3D and AFLW-LPFA datasets. Following the protocol of~\cite{3ddfa}, we randomly select 696 images from AFLW2000-3D, whose absolute yaw angles with small, medium and large values are 1/3 each. 68 points of AFLW2000-3D and 34 points of AFLW-LPFA are selected for evaluation. As illustrated in Table~\ref{tab:1}, our method surpasses the previous methods by a large margin. Specifically, our performance outperforms PRN more than 9\% both on AFLW2000-3D and AFLW-LPFA. This verifies that our method has good face alignment performance even in the case of large face pose. 
\begin{table*}[ht]
\centering
\begin{tabular}{|c|c|c|c|c|c|}
\hline
\multicolumn{1}{|l|}{} & \multicolumn{4}{c|}{AFlW2000-3D}                              & \multicolumn{1}{l|}{AFLW-LFPA} \\ \hline
Method                 & o to 30       & 30 to 60      & 60 to 90      & Mean          & Mean                           \\ \hline
SDM~\cite{cascade2}                   & 3.67          & 4.94          & 9.67          & 6.12          & -                              \\ \hline
3DDFA~\cite{3ddfa}                  & 3.78          & 4.54          & 7.93          & 5.42          & -                              \\ \hline
3DDFA+SDM~\cite{3ddfa}              & 3.43          & 4.24          & 7.17          & 4.94          & -                              \\ \hline
Yu \etal  ~\cite{yu}                 & 3.62             & 6.06             & 9.56             & -             & -                           \\ \hline
3DSTN~\cite{3dtsn}                  & 3.15             & 4.33             & 5.98             & 4.49          & -                           \\ \hline
DeFA~\cite{defa}                  & -             & -             & -             & 4.50          & 3.86                           \\ \hline
PRN~\cite{prn}                   & 2.75          & 3.51          & 4.61          & 3.62          & 2.93                           \\ \hline
Ours                   & \textbf{2.44} & \textbf{3.26} & \textbf{4.35} & \textbf{3.35} & \textbf{2.65}                  \\ \hline

\end{tabular}
\caption{Performance comparison (NME) between our method and other state-of-the-art methods on AFLW2000-3D and AFLW-LPFA benchmarks. The first best result in each category is highlighted in bold, the lower is the better.} 
\label{tab:1}
\end{table*}

\begin{figure}[t]
\centering
\includegraphics[width=3in]{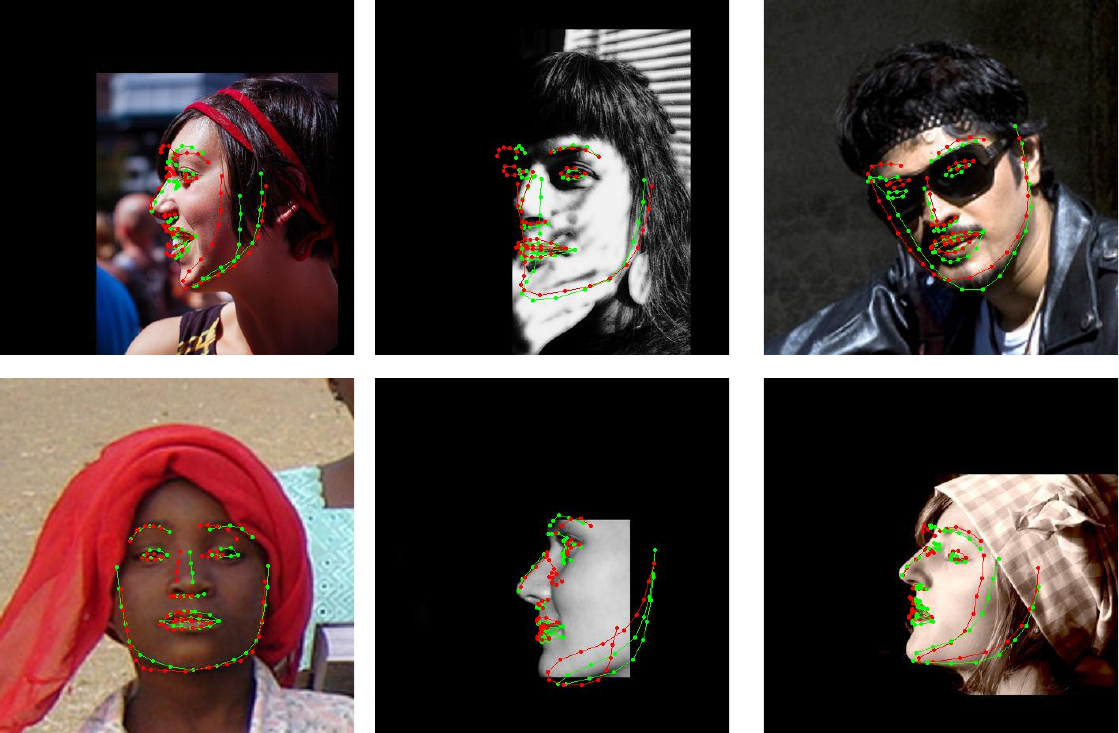}
\caption{Examples from AFLW2000-3D. Our face alignment results are more accurate than the ground truth in some cases. Red is the ground truth, green is our predicted landmarks.}
\label{fig:8}
\end{figure}

Some face alignment examples of AFLW2000-3D are presented in Fig.~\ref{fig:8}, we find that in some cases, our predicted landmarks are more accurate than the ground truth. This is due to that the ground truth points of AFLW2000-3D are generated by the semi-automatic annotation pipeline of~\cite{3ddfa} rather than manual annotation. This phenomenon shows the high accuracy of our method in locating landmarks of faces.

\subsubsection{3D face reconstruction}
\begin{figure}[t]
\centering
\includegraphics[width=3in]{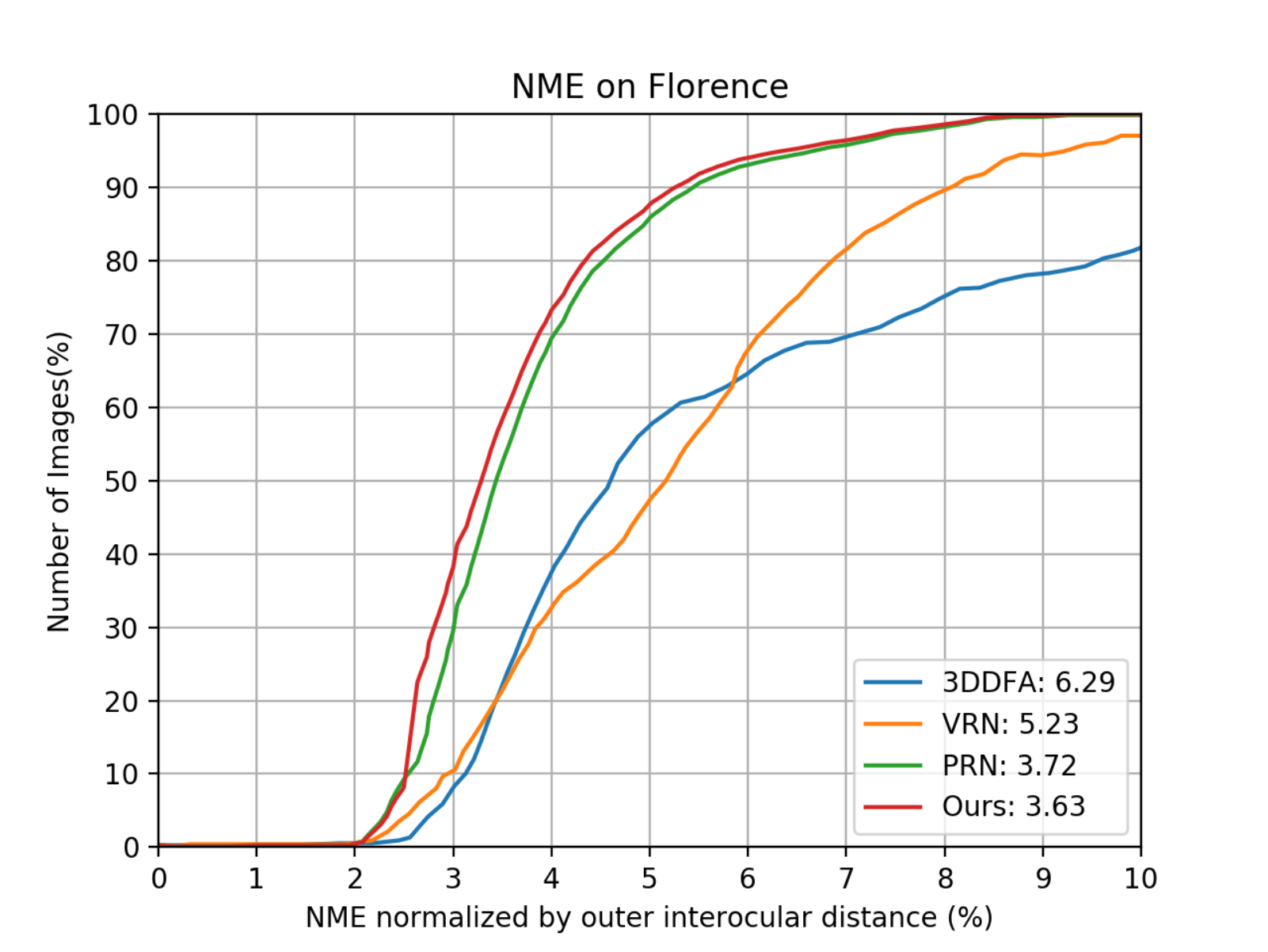}
\caption{Cumulative Errors Distribution (CED) curves of 19K points face reconstruction on orence.}
\label{fig:9}
\end{figure}

\begin{figure}[t]
\centering
\includegraphics[width=3.2in]{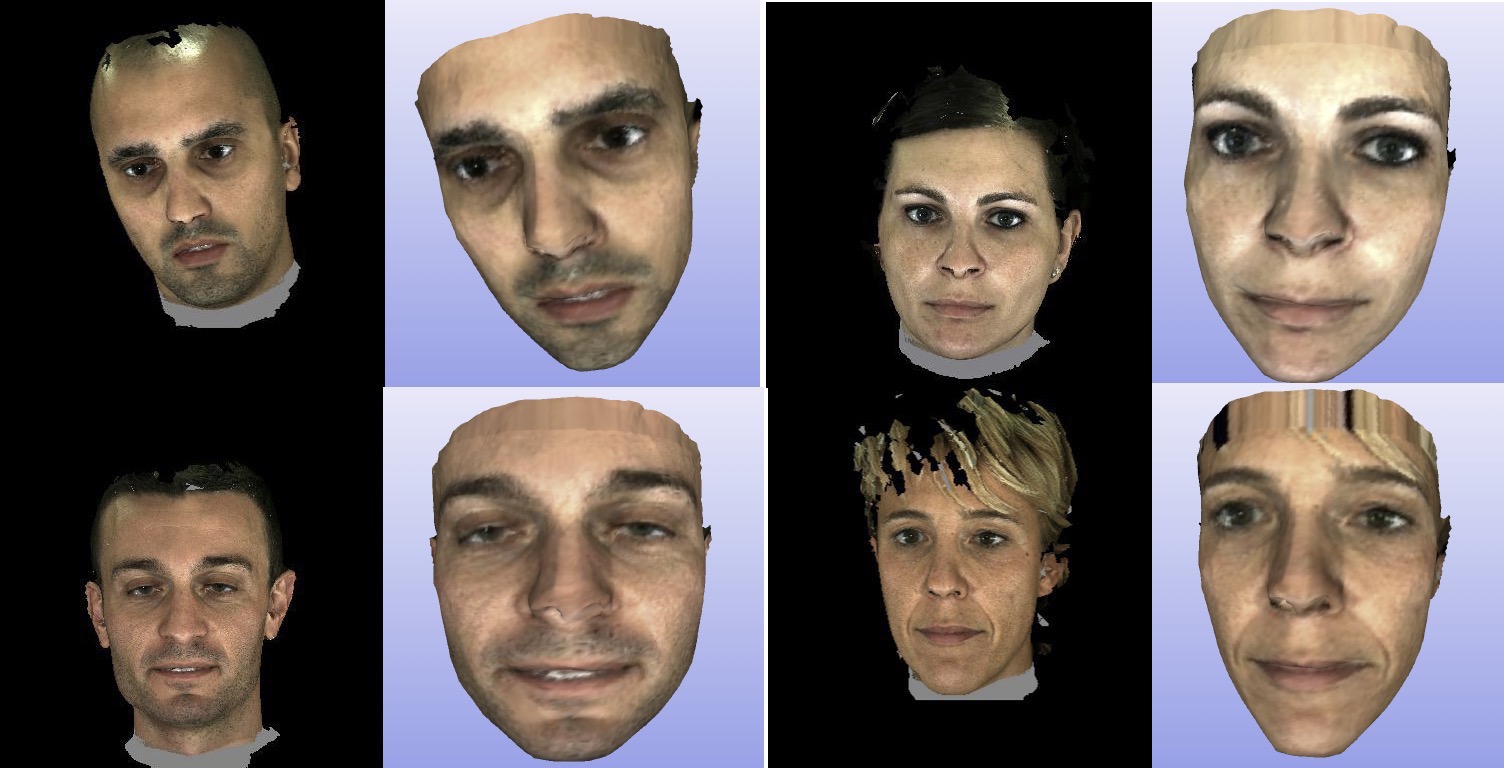}
\caption{Examples of our reconstructed 3D faces  from Florence.}
\label{fig:10}
\end{figure}

In addition to face alignment, we also conduct experiments on 3D face reconstruction task. In this subsection, we investigate the face reconstruction performance of our method on Florence dataset~\cite{florence}. Several state-of-the-art methods are chosen for comparison, including 3DDFA~\cite{3ddfa}, VRN~\cite{vrn} and PRN~\cite{prn}. Following the experimental setting of~\cite{vrn}, we render the testing images of Florence with different poses, the details have been introduced in Section 4.2. 
Before input to the network, the image are cropped using the bounding box computed from the ground truth point cloud. We select the most common 19K points of all compared methods to perform evaluation. As the outputs of different methods are not aligned, we follow~\cite{prn} to utilize Iterative Closest Points algorithm to find the nearest points between the network output and the ground truth point cloud. After aligning the generated point cloud, Mean Squared Error normalized by outer interocular distance of 3D coordinates is adopted as the evaluation metric. The quantitative result is shown in Fig.~\ref{fig:9}, our performance is slightly better than PRN. This is due to the fact that the labeled 3D face meshes of our training data 300W-LP are from 3DMM fitting by~\cite{3ddfa}, while the 3d faces of Florence are acquired from a structured-light scanning system~\cite{florence}, there exists a large gap between the two kinds of mesh annotations. So our model trained on 300W-LP does not bring much performance improvement on Florence data. 
In spite of this, our method still achieve a good correspondence from 2D images to the 3D face meshes, some examples are shown in Fig.~\ref{fig:10}. As is illustrated, the face shape and expression details are well captured by our method.

\begin{figure*}[t]
\centering
\includegraphics[width=6.5in]{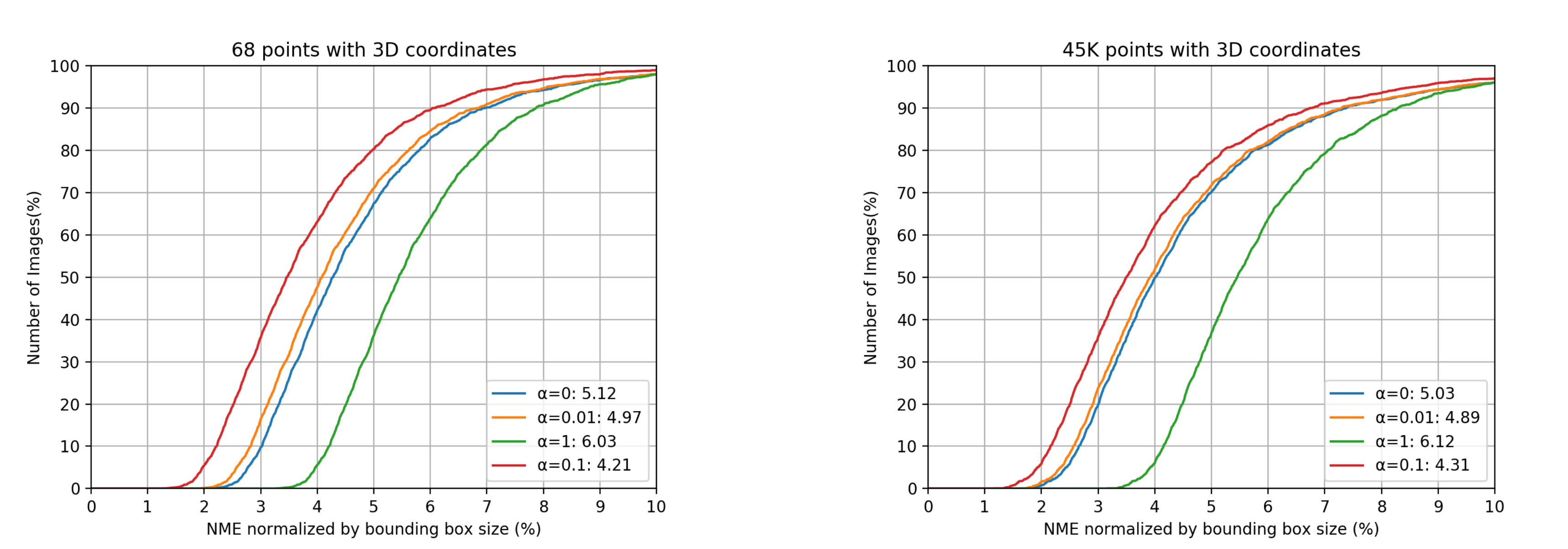}
\caption{Influence of the smooth loss on 3D face alignment performance. The left is performance with 68 points and the right is performance with 45K points.}
\label{fig:11}
\end{figure*}

\begin{figure*}[t]
\centering
\includegraphics[width=6in]{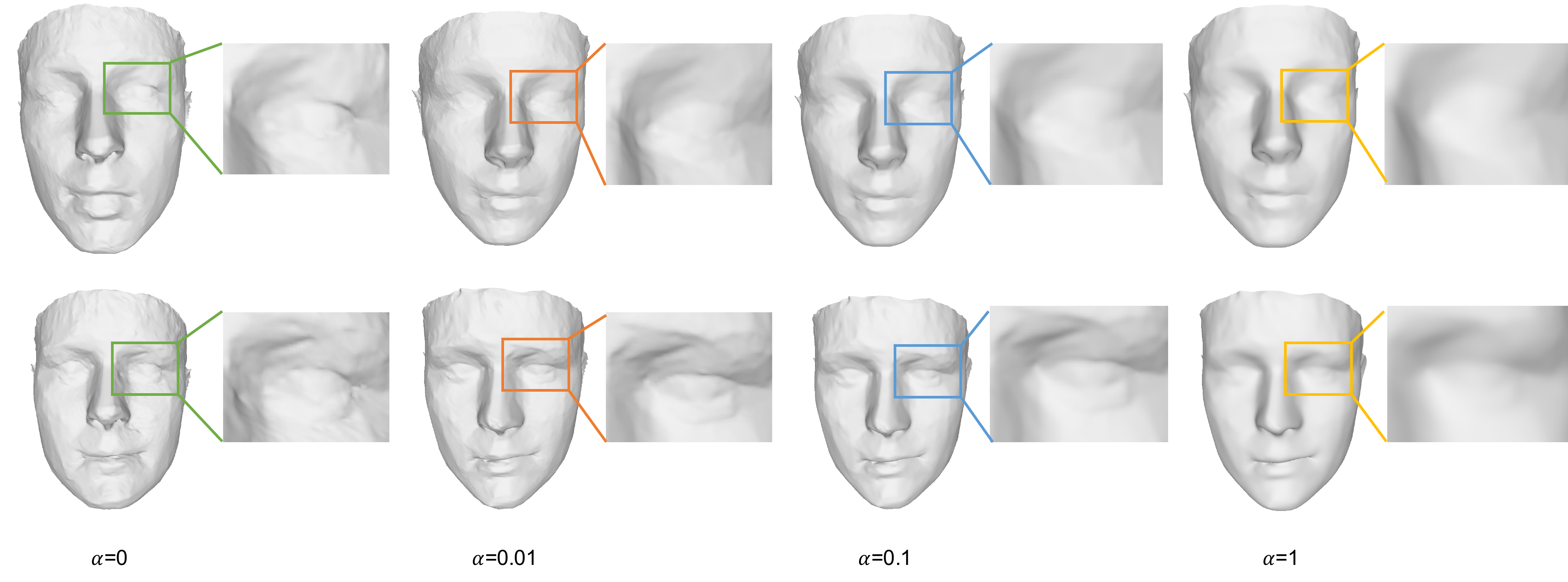}
\caption{Examples of generated 3D faces with different smoothness. The eye area is zoomed in for better view.}
\label{fig:12}
\end{figure*}

\subsubsection{Ablation study}
In this part, we analyze the effectiveness of smooth loss and present quantitative and qualitative results of different weights $\alpha$. Fig.~\ref{fig:11} shows the quantitative face alignment performance comparisons on AFLW2000-3D when $\alpha$ is set to 0, 0.01, 0.1 and 1.0. The corresponding qualitative results are presented in Fig.~\ref{fig:12}. As is illustrated, 0.1 is an appropriate weight which can well balance the the performance and smoothness of the generated 3D faces. This can be interpreted as the following fact, when the value of $\alpha$ is small (0 or 0.01), the generation of 3D face mainly depends on the guidance of $\mathcal{L}_1$, so that the generated surfaces are not smooth. When it becomes large ($\alpha$=1), the smooth loss $\mathcal{L}_{smooth}$ seriously affects the dominant role of $\mathcal{L}_1$ in training procedure. As a result, over-smoothing 3D faces are generated. In contrast, setting $\alpha$ as 0.1 is an appropriate option, which can not only achieve better performance, but also improve the visual quality of 3D face.

\section{Conclusions}
In this paper, we propose graph convolution networks to solve the problem of 3D dense face alignment. Our network captures coarse-to-fine features of face mesh in a hierarchical manner and generate 3D face coordinates directly. Extensive experiments show that our approach gains a superior performance both on face alignment and 3D face reconstruction tasks over other state-of-the-art methods.

{\small
\bibliographystyle{ieee}
\bibliography{egbib}
}

\end{document}